\documentclass[journal]{IEEEtran}

\usepackage{cite}
\usepackage{url}
\usepackage[pdftex]{graphicx}
\graphicspath{{./Figures/}{./Figures/Simulations/}{./Figures/Experiment/}{./Figures/Scheduling_Example/}}
\DeclareGraphicsExtensions{.pdf,.jpeg,.png,.eps}
  
\usepackage[cmex10]{amsmath}
\usepackage{amssymb}
\usepackage{amsthm}
\usepackage{amsfonts}
\usepackage{mathtools}
\usepackage{algorithm}
\usepackage{algorithmic}
\usepackage{array}

\ifCLASSOPTIONcompsoc
  \usepackage[caption=false,font=normalsize,labelfont=sf,textfont=sf]{subfig}
\else
  \usepackage[caption=false,font=footnotesize]{subfig}
\fi

\usepackage{blindtext}
\usepackage{color}
\usepackage{comment}
\usepackage{balance}
\usepackage{enumerate}
\usepackage{epstopdf}
\usepackage{soul}

\newtheorem{theorem}{Theorem}[section]

\newtheorem{problem}[theorem]{Problem}

\begin{document}

\title{Cooperative Periodic Coverage with Collision Avoidance}

\author{Jos\'e~Manuel~Palacios-Gas\'os,~\IEEEmembership{Student Member,~IEEE,}
		Eduardo~Montijano,~\IEEEmembership{Member,~IEEE,}
		Carlos~Sag\"u\'es,~\IEEEmembership{Senior Member,~IEEE,}
		and Sergio~Llorente
\thanks{J.M. Palacios-Gas\'os, E. Montijano and C. Sag\"u\'es are with Instituto de Investigaci\'{o}n en Ingenier\'{\i}a de Arag\'{o}n (I3A), Universidad de Zaragoza, Zaragoza, Spain.
{\tt\small \{jmpala, emonti, csagues\}@unizar.es}}
\thanks{S. Llorente is with Research and Development Department, Induction Technology, Product Division Cookers, BSH Home Appliances Group, 50016, Zaragoza, Spain.
{\tt\small{sergio.llorente@bshg.com}}}
\thanks{This work was partially supported by projects RTC-2014-1847-6 of Retos-Colaboraci\'on, DPI2015-69376-R from Ministerio de Econom\'ia y Competitividad/Uni\'on Europea, DGA T04-FSE, CUD2016-17 and DGA Scholarship C076/2014, partially funded by European Social Fund.}
 \thanks{\textcolor{red}{This is the accepted version of the manuscript: J. M. Palacios-Gasós, E. Montijano, C. Sagüés and S. Llorente, ``Cooperative Periodic Coverage With Collision Avoidance," in IEEE Transactions on Control Systems Technology, vol. 27, no. 4, pp. 1411-1422, July 2019, doi: 10.1109/TCST.2018.2823278.
\textbf{Please cite the publisher's version}. For the publisher's version and full citation details see:\\
\protect\url{https://doi.org/10.1109/TCST.2018.2823278}. 
}}

 \thanks{© 2019 IEEE.  Personal use of this material is permitted.  Permission from IEEE must be obtained for all other uses, in any current or future media, including reprinting/republishing this material for advertising or promotional purposes, creating new collective works, for resale or redistribution to servers or lists, or reuse of any copyrighted component of this work in other works.}
}

\maketitle

\begin{abstract}
In this paper we propose a periodic solution to the problem of persistently covering a finite set of interest points with a group of autonomous mobile agents. These agents visit periodically the points and spend some time carrying out the coverage task, which we call coverage time.
Since this periodic persistent coverage problem is NP-hard, we split it into three subproblems to counteract its complexity.
In the first place, we plan individual closed paths for the agents to cover all the points. Second, we formulate a quadratically constrained linear program to find the optimal coverage times and actions that satisfy the coverage objective. Finally, we join together the individual plans of the agents in a periodic team plan by obtaining a schedule that guarantees collision avoidance. To this end, we solve a mixed integer linear program that minimizes the time in which two or more agents move at the same time.
Eventually, we apply the proposed solution to an induction hob with mobile inductors for a domestic heating application and show its performance with experiments on a real prototype.
\end{abstract}

\section{Introduction}
\label{sec:introduction}

In recent years, the development of autonomous mobile agents has experienced an important growth and has provided them the capabilities to carry out a great variety of tasks while being affordable.
This has motivated an increasing research interest in multi-agent systems that are capable of accomplishing tasks as a team that a single agent cannot~\cite{khamis2015mrtaskassignment,hussein2007effective}.
A particularly interesting problem in this context is that of persistent coverage, which finds applications in many fields such as cleaning~\cite{kakalis2008oilspill}, environmental monitoring~\cite{smith2011persistent,jadaliha2013envmonitoringaquatic} or aerial delivery~\cite{mitchell2015percovstochastic}.
The objective of persistent coverage is to maintain covered over time a dynamic environment in which the coverage level persistently deteriorates.
Coverage can be seen as a measure of how dirty is the environment in a cleaning application, how hot in a heating one or how well observed in monitoring. As time goes by, the environment gets dirty, cools down or the accuracy of the observations degrades, respectively.
The agents must spend some time at each point of the environment and keep moving to maintain the coverage of the entire environment at a desired level.
This is the main difference with more traditional coverage approaches~\cite{cortes2004coverage}.

\subsection{Related Work}

Different types of solutions, which can be indefinitely applied~\cite{hokayem2007cdc}, have been proposed.
The first type are controller-based approaches as~\cite{palacios2016control}, where a gradient descent method together with an assignment of objectives is proposed.
In~\cite{nigam2012uavsperssurv} a control policy decides to which cell move next to minimize the maximum time between visits to all cells and Branch and Bound is used in~\cite{palacios2016branchandbound} to find the optimal control inputs of the agents for a finite prediction horizon.

The second kind of proposals intend to plan closed paths that the agents repeatedly follow. The objective is to minimize some coverage metric while periodically visiting all the points of the environment.
To minimize the estimation uncertainty, a parametric optimization is formulated in~\cite{lin20132D}, Rapidly-exploring Random Trees are used in~\cite{lan2013planperstraject} and Rapidly-exploring Random Cycles in~\cite{lan2016rrc}.
In~\cite{portugal2014routes}, they aim to minimize the maximum time that points remain unvisited by planning paths through the vertices of a graph and~\cite{song2013awareness} guarantees that full awareness is eventually reached in some interest points.
In these proposals, the time that each agent needs to perform the coverage task at each point is not considered and the movements have priority over the coverage of particular points.

In discrete environments, where only a finite set of points requires coverage, approaches are related to Task Assignment~\cite{korsah2013mrtaskalloctaxonomy} and Vehicle Routing Problems (VRP) as addressed in operational research~\cite{schupbach2013adaptiverouting}.
Since VRP are known to be NP-hard, three different types of solutions appear in the literature: based on heuristics, formulated as Mixed Integer Programs (MIP), although they suffer from complexity for large problems, or a combination of both.
In~\cite{kim2013schedulingUAV} a Mixed-Integer Linear Program (MILP) and a genetic algorithm are compared for the scheduling of mission trajectories.
To find a closed path to visit all the points of the environment with a single UAV, heuristic solutions are presented with refueling depots~\cite{sundar2014routingdepots} and with revisit constraints~\cite{fargeas2013revisitconstraints}.
In~\cite{mitchell2015fuelconstrained}, they seek the optimal routing strategy using a linear MIP that allows a fleet of vehicles to periodically visit a set of targets while trying to minimize the energy consumption.
Similarly, in~\cite{mitchell2015percovstochastic} they maximize the frequency of task completion allowing online calculation of task costs to avoid energy depletion.
These approaches assume that only one agent covers each point.
In~\cite{yu2016orientpermonit} a utility metric is maximized through a quadratic MIP to find the best tours for the agents when correlation between the points of interest is considered.

Most of the aforementioned works assume that the times required to complete the coverage task at each point are known or do not consider them. The latter is acceptable in monitoring or surveillance applications~\cite{nigam2014review}, since the information gathering can be considered instantaneous, but the same does not hold in problems such as heating or watering, where the coverage action of the agent requires some time. Moreover, the possibility of collisions between agents along their paths is usually not addressed.

The problem of calculating the times that the agents have to spend at each point to satisfy the coverage objective has not been deeply studied in the literature.
In~\cite{song2013awareness} and~\cite{smith2012persistenttask} velocity controllers are calculated to spend more time covering the points where the environment changes quickly.
In our previous work~\cite{palacios2016times}, on which this work builds, the optimal coverage times for each pair agent-point are calculated when the actions that the agents apply are fixed.
Without a coverage objective to accomplish, in~\cite{yu2015optimschedsensingrobot} a single agent obtains a reward depending on the time that it covers each point and a path is calculated to obtain the maximum reward with a limited amount of fuel.

The problem of collision avoidance is considered in other multi-agent works by scheduling the agent trajectories.
This is usually solved by introducing time delays at the beginning of the paths~\cite{wang2015collavoiddelays}. In~\cite{turpin2014goaltrajectmr} they also assign goals to interchangeable agents to minimize the maximum cost over all trajectories.

\subsection{Contributions}

In this paper we propose a solution to persistently cover a finite set of points with a team of agents. These agents must spend some time at each point to improve its coverage.
We find a periodic strategy for the team, composed of individual periodic paths, coverage times and actions, that guarantees the satisfaction of the coverage objective for all the points, optimizes the actions of the agents and avoids collisions between them.

Our main contribution is a complete solution to the problem using a divide-and-conquer strategy.
We separate the problem into three subproblems:
(i) planning individual, closed paths for the agents that cover all the points of the environment;
(ii) calculating the coverage time and coverage action of each agent at each point of its path to satisfy the coverage objective; and
(iii) scheduling the start of the agent paths to obtain a team plan in which collisions are avoided.

As mentioned before, the complete problem is hard to solve at once because the computational complexity grows exponentially with the number of agents and points.
The divide-and-conquer methodology counteracts this growth and allows us to find the optimal solution for each subproblem independently. Moreover, although the final solution to the entire problem may not be the optimal, this is a conceptually simple and piecewise optimal way of calculating a solution to the problem. In fact, this is the first attempt to calculate simultaneously the paths and the coverage times and actions, that allows the collaboration of several agents at the same point and guarantees collision avoidance.

In the paper we do not elaborate on subproblem (i), since it reduces to solve an instance of the Traveling Salesman Problem (TSP) for which there are state-of-the-art solutions.
Regarding subproblems (ii) and (iii), our contributions in this work are:
\begin{itemize}
\item A generalization of our previous work~\cite{palacios2016times} that includes the coverage actions with the coverage times in subproblem (ii). It allows us to optimize the coverage satisfying the objective and minimizing the instantaneous actions of the agents.
\item A procedure to schedule the agent paths that provides collision avoidance guarantees and minimizes simultaneous movements of the agents.
\item An application of the solution to the problem of domestic induction heating with mobile inductors, that is an open problem which offers a significant opportunity to increase the flexibility of domestic hobs at a moderate increase of cost.
\item Real experiments with a prototype for this application.
\end{itemize}

The remainder of the paper is structured as follows.
In Section~\ref{sec:formulation} we introduce the problem formulation and an overview of the proposed solution.
We calculate the optimal coverage times and actions in Section~\ref{sec:opttimespower} and face the team plan scheduling problem in Section~\ref{sec:scheduling}.
Finally, we present simulation results in Section~\ref{sec:simulations}, the application to domestic induction heating with experimental results in Section~\ref{sec:experiments} and conclusions in Section~\ref{sec:conclusions}.

\section{Problem Formulation and Solution Overview}
\label{sec:formulation}

Let $\mathcal{Q} = \{\mathbf{q}_1,\dots,\mathbf{q}_Q\}$
be a finite set of $Q$ points of interest that must be covered.
At each point, we define the coverage level with a scalar field, $Z(\mathbf{q},t) \geq 0$.
The objective is to maintain a desired level $Z^*(\mathbf{q},t) \geq 0$ by providing a certain coverage action $P^*(\mathbf{q},t) \geq 0$ over time, not necessarily the same for all the points.

This is achieved using a team of $I \in \mathbb{N}$ mobile agents $\mathcal{I} = \{i_1,\dots,i_I\}$ of radius $r_i, \, i\in\mathcal{I}$. They are capable of increasing the level at the points in which they are located according to a \emph{production function}, $P_i(\mathbf{q},t)$:
\begin{equation*}
 P_i(\mathbf{q},t) = \left\{
  \begin{aligned}
    & 0, & \text{if} \; \mathbf{p}_i(t) \neq \mathbf{q}, \\
    & \rho_i(\mathbf{q}), & \text{if} \; \mathbf{p}_i(t) = \mathbf{q},
  \end{aligned}
  \right.
\end{equation*}
where $\mathbf{p}_i(t)$ is the position of the agent at time $t \geq 0$ and $0 \leq \rho_i(\mathbf{q}) \leq \rho_i^{\max}(\mathbf{q})$ is the \emph{coverage action} of the agent, which can also be controlled. The maximum production $\rho_i^{\max}(\mathbf{q})$ can be different for each agent-point pair.

Each agent may not be capable of reaching all the points due to physical constraints.
The subset $\mathcal{Q}_i \subseteq \mathcal{Q}$ represents the \emph{reachable set} of points for agent $i,$
with the team satisfying $\cup_{i \in \mathcal{I}} \mathcal{Q}_i = \mathcal{Q}$.
Reciprocally, each point $\mathbf{q} \in \mathcal{Q}$ can be covered by a set of visiting agents, $\mathcal{I}_\mathbf{q} \subseteq \mathcal{I}$.

In this context, the coverage provided at each point can be calculated as
\begin{equation}
Z(\mathbf{q},t) = \int_0^t \sum\nolimits_{i \in \mathcal{I}_\mathbf{q}} P_i(\mathbf{q},\tau) d\tau,
\label{Eq:Z}
\end{equation}
assuming $Z(\mathbf{q},0) = 0$ as initial condition, and the desired coverage as
\begin{equation}
Z^*(\mathbf{q},t) = \int_0^t P^*(\mathbf{q},\tau) d\tau.
\label{Eq:Z_obj}
\end{equation}
A solution that maintains all the time $Z(\mathbf{q},\tau) = Z^*(\mathbf{q},\tau)$ requires at least as many agents as points and does not exist if the number of agents is lower than the number of points.
Since this is usually the case, an optimization metric is defined to try to maintain $Z(\mathbf{q},\tau)$ as close as possible to $Z^*(\mathbf{q},\tau)$ over time.
Some optimality criteria could be to minimize over time the quadratic difference between the required and the provided coverage, i.e., minimize
$ \int_0^t \sum_{\mathbf{q}\in\mathcal{Q}} \big(Z^*(\mathbf{q},\tau) - Z(\mathbf{q},\tau) \big)^2 d\tau,$
or to minimize the maximum difference between $Z(\mathbf{q},t)$ and $Z^*(\mathbf{q},t)$ over time, i.e., minimize $\max\big(Z^*(\mathbf{q},t)-Z(\mathbf{q},t)\big)$.
However, trying to minimize globally a function associated with these metrics results in a problem that has been proven to be NP-hard.
For these reason, we seek to guarantee that the desired coverage is provided periodically with period $T$.
A periodic approach, though inherently suboptimal with respect to such metrics and still NP-hard, allows us to guarantee that all the points receive on average the coverage that they require, with a repetitive strategy that can be calculated in advance.
In this scenario, the periodic objective requires that (i) at some time $t \leq T$ the coverage reaches $Z(\mathbf{q},t) = Z^*(\mathbf{q},t)$, and (ii) from that time on, $Z(\mathbf{q},t + kT) = Z^*(\mathbf{q}, t + kT), \, \forall \, k \in \mathbb{Z}$. Equivalently,
\begin{multline*}
Z(\mathbf{q},t+kT) - Z\big(\mathbf{q},t+(k-1) \, T\big) = \\
Z^*(\mathbf{q},t+kT) - Z^*\big(\mathbf{q},t+(k-1) \, T\big).
\end{multline*}
We can assume that $P^*(\mathbf{q}) \equiv P^*(\mathbf{q},t)$ is constant over time, or at least periodic, since the rate of change of $P^*(\mathbf{q})$ is usually much bigger than the period. Thus, the problem can be considered stationary between changes.
Then, the right term is equal to $P^*(\mathbf{q},\tau) \, T$ according to~\eqref{Eq:Z_obj} and introducing~\eqref{Eq:Z} the objective becomes
\begin{equation*}
\int_{t+(k-1)T}^{t+kT} \sum\nolimits_{i \in \mathcal{I}_{\mathbf{q}}} P_i(\mathbf{q},\tau) d\tau = P^*(\mathbf{q}) \, T.
\end{equation*}
In order to satisfy this objective, the agents follow a periodic path, $\Gamma_i$, between reachable points, i.e., an ordered subset of $\mathcal{Q}_i$.
Along these paths, the agents spend some time, namely the \emph{coverage time} $\theta_{i,\Gamma_i(j)}$, covering each point. 
These times are normalized by the period, i.e., they represent the fraction of the period that a robot is covering a point and allow us to formulate the problem independently of the actual value of the period. With this normalization, the periodic coverage objective for each point $\mathbf{q}$ of the environment can be stated as
\begin{equation*}
 \sum\nolimits_{i \in \mathcal{I}_\mathbf{q}} \rho_{i,\Gamma_i(j)} \, \theta_{i,\Gamma_i(j)} = P^*(\mathbf{q}).
\label{Eq:PeriodicObjFull}
\end{equation*}
With a little abuse of notation we refer to $\mathbf{q}$ as the j-th point in the path of agent $i$, i.e., $\Gamma_i(j) \equiv \mathbf{q}$, although $j$ may be different for each agent $i$.
In the following we refer to it only with subindex $j$ for simplicity.
Thus, the objective of the problem is to find the paths $\Gamma_i$, coverage times $\theta_{i,j}$ and productions $\rho_{i,j}$ that satisfy
\begin{equation}
 \sum\nolimits_{i \in \mathcal{I}_\mathbf{q}} \rho_{i,j} \, \theta_{i,j} = P^*(\mathbf{q})
\label{Eq:PeriodicObj}
\end{equation}
and do not lead to a collision.

Each point is visited only once per period by the agent and each visit has associated two additional times: the \emph{arrival time}, $a_{i,j}$, the instant in which the agent arrives and starts covering a point, and the \emph{departure time}, $d_{i,j}$, the instant in which the agent stops covering the point and leaves towards the next one.
The total (normalized) moving time, i.e., the time that agent $i$ would need to traverse $\Gamma_i$ without making any stops, is $\theta^m_i$.
Note that the periodic plans can be repeated as long as all $\mathcal{Q}_i$, $\mathcal{I}_\mathbf{q}$ and $P^*(\mathbf{q})$ remain constant.
Also note that they are independent of the value of the period and that such value can be chosen differently depending on the application.

The solution that we propose to this problem follows a divide-and-conquer strategy, as shown in Alg.~\ref{Alg:Solution}. We separate the problem into three subproblems (Steps~\ref{AlgSt:Paths}, ~\ref{AlgSt:Times} and~\ref{AlgSt:Sched}) to reduce its complexity and make it tractable. 
Even though it does not guarantee global optimality, this allows us to solve efficiently the three subproblems and guarantee periodic coverage. In fact, the second and third subproblems can be solved in an optimal manner and for the first one, that has been deeply studied in the literature, there are many efficient methods with guaranteed closeness to the optimum.

\begin{algorithm}[h!]
\caption{Solution Overview}
\label{Alg:Solution}
\begin{algorithmic}[1]
\STATE Plan initial paths. \label{AlgSt:Paths}
\STATE Calculate optimal coverage times and actions. \label{AlgSt:Times}
\STATE Shorten paths and recalculate times and actions.
\STATE Schedule the path starts to avoid collisions. \label{AlgSt:Sched}
\end{algorithmic}
\end{algorithm}

In the first step, we find the initial path for each robot. This path is the shortest, closed one that visits all the reachable points of the robot, $\mathcal{Q}_i$. To do so, we solve a simple Traveling Salesman Problem (TSP)~\cite{christofides1976heuristictsp} using an order-first split-second approach~\cite{prins2014order1split2}. We do not elaborate further on this step since many solutions have been proposed and analyzed in the literature.
For the second step, we pose a quadratically constrained linear program with the restrictions of satisfying the periodic coverage objective from~\eqref{Eq:PeriodicObj}.
The algorithm to shorten the paths can be seen in our previous work~\cite{palacios2016times}.
In the last step, we calculate a schedule that allows us to include the individual periodic plans of the agents in a periodic team plan in which collision avoidance is guaranteed. We formulate this scheduling as a MILP to minimize the time in which two or more agents are moving at the same time while satisfying collision avoidance restrictions. The problem could be limited to satisfy them but it is worth to optimize another criteria at the same time.
Actually, the restrictions might be included in the problem of the times and production, but this would exponentially increase the complexity of the problem and would still not guarantee the global optimum to be found.

In the following sections we explain in detail the two main steps of our solution, namely Steps \ref{AlgSt:Times} and \ref{AlgSt:Sched}.

\section{Optimal Times and Productions}
\label{sec:opttimespower}

The periodic coverage objective requires the calculation of the coverage times and coverage actions to be guaranteed, with the periodic paths that the agents travel.
The problem of calculating these times and productions reduces to find the times $\theta_{i,j}$ and productions $\rho_{i,j}$ that comply with~\eqref{Eq:PeriodicObj} and with the periodicity of the system.
To this end, we consider a cost function on the times and the productions, $f(\theta_{i,j}, \rho_{i,j})$, and the resulting quadratically constrained program is
\begin{subequations}
\label{Eq:QuadProgram}
\begin{align}
& \underset{\theta_{i,j}, \rho_{i,j}}{\text{minimize}} & 
& f(\theta_{i,j}, \rho_{i,j}) & \nonumber\\
& \text{subject to} & & \sum_{i \in \mathcal{I}_\mathbf{q}} \rho_{i,j} \, \theta_{i,j} = P^*(\mathbf{q}), & \forall \, \mathbf{q} \in \mathcal{Q},
\label{Subeq:QuadProgPeriodicObj} \\
& & & \sum_{\mathbf{q} \in \mathcal{Q}_i} \theta_{i,j} \leq 1 - \theta^m_i, & \forall \, i \in \mathcal{I}, \label{Subeq:QuadProgPeriodRobots} \\
& & & \sum_{i \in \mathcal{I}_\mathbf{q}} \theta_{i,j} \leq 1, & \forall \, \mathbf{q} \in \mathcal{Q}. \label{Subeq:QuadProgPeriodPoints}
\end{align}
\end{subequations}

Eq.~\eqref{Subeq:QuadProgPeriodicObj} is the quadratic restriction on the coverage objective. of a linear program that has already been solved distributedly~\cite{richert2015distributedlp} for multi-robot systems~\cite{montijano2014formationdistopt}.
The set of equations~\eqref{Subeq:QuadProgPeriodRobots} imposes that for each agent the time spent covering the points plus the time to move along the path must be lower than the period, and the set~\eqref{Subeq:QuadProgPeriodPoints} represents that each point cannot be covered for more time than the period.
This restriction is only needed in the case that it is not allowed that two agents cover the same point at the same time because otherwise it would be impossible to avoid collisions between agents.
The selection of the cost function is entirely dependent on the performance expected from the particular application.
For this reason, in Section~\ref{sec:simulations} we explore and analyze different linear cost functions that are appropriate for many applications.

This optimization problem is a generalization of our previous approach~\cite{palacios2016times} and includes the productions of the agents at each point as variables of the problem.
Nevertheless, the sufficient conditions on the existence of solution presented in~\cite{palacios2016times} are still valid.
These conditions allow us to know a priori if the agents are capable of satisfying the coverage objective.

A particularity of the solutions obtained for~\eqref{Eq:QuadProgram} is that some of the times $\theta_{i,j}$ may be equal to zero.
This implies that agent $i$ is not required to cover point $j$.
Thus, shorter paths can be followed only through points with coverage times greater than zero, reducing the moving times $\theta_i^m$ and, therefore, the total time required for the coverage of the environment.

In our previous work~\cite{palacios2016times} we proposed an iterative algorithm to shorten the paths based on this property that links the first two subproblems. At each iteration, the algorithm optimizes the agent paths using the results of the previous optimal coverage times and productions. Then, the times and productions are optimized again for the new refined paths.
This makes the combination of both to iteratively improve the global solution achievable with this periodic approach.

\section{Team Plan Scheduling}
\label{sec:scheduling}

We devote this section to the calculation of a scheduling for the start of the periodic paths of the agents. It avoids collisions while covering a point and while moving between points.
The intuitive idea of the scheduling is to shift the individual plans of the agents in time to obtain a collision-free plan for the entire team, that is, to find a time, $0 \leq \varphi_i < 1$, for each agent $i \in \{2,\dots,I\}$ such that the execution of its periodic plan shifted by this time leads to no collision with the others.
We refer the initial times to the beginning of the path of agent $i=1$, i.e., we fix $\varphi_1 = 0$, and the paths remain the same through the scheduling.

\begin{figure}[!ht]
        \centering
        \subfloat[Individual plans before scheduling.]{
                \includegraphics[width=0.23\textwidth]{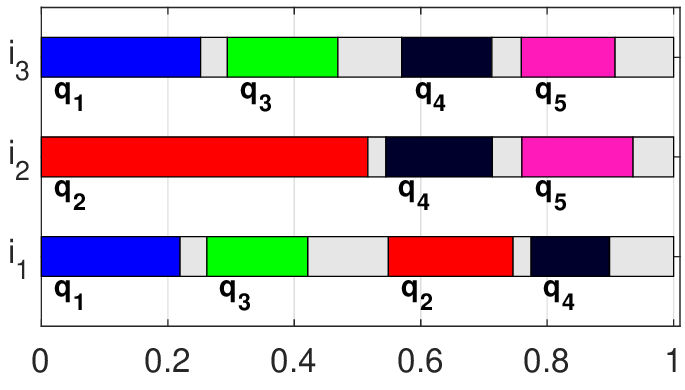}
                \label{Subfig:NoScheduling}
                }
                \hfil
        \subfloat[Team plan after scheduling.]{
             \includegraphics[width=0.23\textwidth]{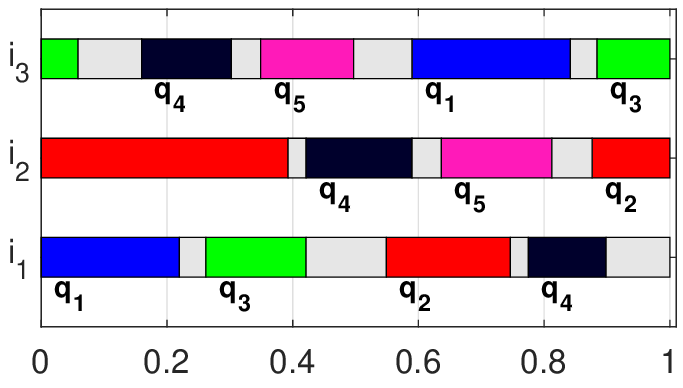}
                \label{Subfig:Scheduling}
                }
                
        \subfloat[Agent paths.]{
             \includegraphics[width=0.23\textwidth]{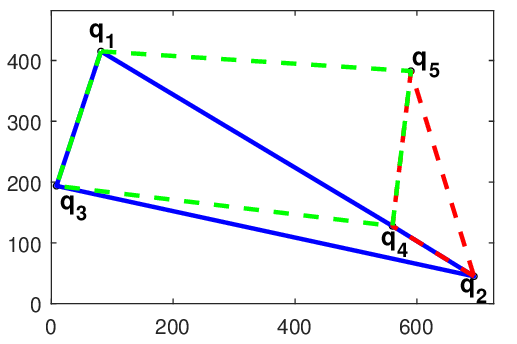}
                \label{Subfig:Paths}
                }
    \caption{Example of scheduling. (a)-(b) Each row represents the plan of an agent. The points $\mathbf{q}_i$ that each agent covers are represented in different colors. The coverage times are depicted by the width of the colored rectangles and the gray rectangles represent the time needed to move between points. (c) Paths followed by the agents $i_1, i_2$ and $i_3$ in blue, red and green, respectively. The radius of the agents is $r_i=5$ units.}
    \label{Fig:SchedulingExample}
\end{figure}

In Fig.~\ref{Fig:SchedulingExample} we show an example of scheduling for $I = 3$ agents of size $r_i =5$ and $Q=5$ points. Fig.~\ref{Subfig:NoScheduling} shows the individual plan of each agent, which includes the path (order in which points are visited) and the coverage times (width of the colored rectangles). The beginning and end of the colored rectangles correspond to the arrival and departure times, $a_{i,j}$ and $d_{i,j}$, respectively, and the gray rectangles represents the time needed to move from one point to the following. It can be seen that some coverages of the same point are overlapped in time and, therefore, the execution of these individual plans leads to collisions. For instance, it happens when $i_1$ and $i_3$ try to cover $\mathbf{q}_1$ or $\mathbf{q}_3$, or when $i_2$ and $i_3$ try to cover $\mathbf{q}_4$ or $\mathbf{q}_5$.
In Fig.~\ref{Subfig:Scheduling}, where the team plan after the scheduling is depicted, it can be seen that the individual plans of agents $i_2$ and $i_3$ have been shifted an $88 \%$ and a $59 \%$ of the period respectively and that in the resulting plan there are no overlaps between coverages of the same point.
Fig.~\ref{Subfig:Paths} shows the locations of the points and the paths followed by the agents. It can be seen that no collisions occur after the scheduling.

The calculation of the optimal schedule is done by solving a MILP with constraints. In the following subsections we develop the formulation, the restrictions and the cost function, and finally pose the problem and discuss its suitability.

\subsection{Transformation of Individual Times to Team Plan}

The previous step to calculate the schedule that avoids collisions is to refer the individual times of the agents to the team plan. The arrival times can be expressed as follows:
\begin{subequations}
  \label{Eq:ArrTimesSystem}
  \begin{align}
    A_{i,j} & = \varphi_i + a_{i,j}, & \text{if} \; \varphi_i + a_{i,j} \leq 1, \label{Subeq:ArrTimesSystem1} \\
    A_{i,j} & = \varphi_i + a_{i,j} - 1, & \text{if} \; \varphi_i + a_{i,j} > 1,\label{Subeq:ArrTimesSystem2}
  \end{align}
\end{subequations}
for all $j \in \{2,\dots,|\Gamma_i|\}$, $i \in \{2,\dots,I\}$.
Note that~\eqref{Eq:ArrTimesSystem} does not apply for $j=1$ since $a_{i,1} = 0$ and $A_{i,1} = \varphi_i$, and recall that for agent $i=1$ we set $\varphi_1 = 0$ and, therefore, its corresponding times do not need transformation.

To be able to include these times in the MILP formulation, we introduce the binary variables, $c^a_{i,j}$, such that, $c^a_{i,j} = 0$ if $\varphi_i + a_{i,j} \leq 1$ and $c^a_{i,j} = 1$, otherwise.
These variables represent if the arrival times in the team plan are bigger than 1 and they have to be shifted to the left side of the team plan by substracting 1, e.g., the arrival time of $i_2$ to $\mathbf{q}_4$ in Fig.~\ref{Subfig:Scheduling}.

Thus, we can express the arrival times as
\begin{equation*}
A_{i,j} = \varphi_i + a_{i,j} - c^a_{i,j}
\end{equation*}
subject to
\begin{subequations}
\label{Eq:Constr_1_12}
\begin{align}
\varphi_i + a_{i,j} & \leq 1 + R \, c^a_{i,j}, \\
-(\varphi_i + a_{i,j}) & \leq -1 + R \, ( 1 - c^a_{i,j}),
\end{align}
\end{subequations}
obtained with the big number method~\cite{griva2009linear}. This method activates or deactivates the constraints depending on the value of the binary variable with $R$ being a sufficiently big number. In this case, $c^a_{i,j} = 0$ if~\eqref{Subeq:ArrTimesSystem1} has to be used and $c^a_{i,j} = 1$ if~\eqref{Subeq:ArrTimesSystem2} is needed.

The same formulation can be developed for the departure times:
\begin{align*}
D_{i,j} & = \varphi_i + d_{i,j}, & \text{if} \; \varphi_i + d_{i,j} \leq 1, \\
D_{i,j} & = \varphi_i + d_{i,j} - 1, & \text{if} \; \varphi_i + d_{i,j} > 1,
\end{align*}
for all $j \in \{1,\dots,|\Gamma_i|\}$, $i \in \{2,\dots,I\}$.
Introducing the binary variables $c^d_{i,j}$, we have
\begin{equation*}
D_{i,j} = \varphi_i + d_{i,j} - c^d_{i,j}
\end{equation*}
subject to
\begin{align*}
\varphi_i + d_{i,j} & \leq 1 + R \, c^d_{i,j}, \\
-(\varphi_i + d_{i,j}) & \leq -1 + R \ ( 1 - c^d_{i,j}).
\end{align*}

Since the binary variables $c^a_{i,j}$ and $c^d_{i,j}$ are variables of the problem, we include the following restrictions on their values to guarantee that the order of the paths is followed:
\begin{align*}
    c^a_{i,j} - c^d_{i,j} & \leq 0, \quad \forall \, j \in \{2,\dots,|\Gamma_i|\}, \\
    c^d_{i,j} - c^a_{i,j+1} & \leq 0, \quad \forall \, j \in \{1,\dots,|\Gamma_i|-1\}.
\end{align*}

\subsection{Collision Avoidance During Coverage}

A thorough consideration of collisions requires the inclusion of spatio-temporal restrictions in the problem, which implies a discretization of the environment and an analysis of what happens at each point~\cite{wang2015collavoiddelays}.
Nevertheless, for large environments or with many agents or points to cover the problem becomes computationally unaffordable or even intractable.
For this reason, we include the collision avoidance as planning constraints.
Although it is overprotective, it is much lighter in terms of computational cost.

The first situation in which a collision may occur is when two or more agents try to cover the same point at the same time. From the planning perspective, this happens if the coverages of such point by different agents are overlapped in the team plan. We avoid this type of collisions with two groups of constraints. The first group is
\begin{subequations}
\label{Eq:Constr_3_1}
\begin{align}
    A_{i_1,j_1} - A_{i_2,j_2} & \leq R \, (1 - c^{cc}_{i_1.j_1,i_2,j_2}), \label{12a_d} \\
    A_{i_1,j_1} - D_{i_1,j_1} & \leq R \, (1 - c^{cc}_{i_1.j_1,i_2,j_2}), \label{12a_e} \\
    D_{i_1,j_1} - A_{i_2,j_2} & \leq -\varepsilon + R \, (1 - c^{cc}_{i_1.j_1,i_2,j_2}), \label{12a_a}\\
    D_{i_2,j_2} - A_{i_1,j_1} & \leq -\varepsilon + R \, (1 - c^{cc}_{i_1.j_1,i_2,j_2}) \nonumber \\
    & \quad + R (1 - c^d_{i_2,j_2} + c^a_{i_2,j_2}), \label{12a_c}
\end{align}
\end{subequations}
for all $j_1,j_2$ such that $\Gamma_{i_1}(j_1) = \Gamma_{i_2}(j_2)$, where the constant $\varepsilon$ is the minimum separation between departures and arrivals to the same point.
The first restriction, Eq.~\eqref{12a_d}, activates this group if $A_{i_1,j_1} \leq A_{i_2,j_2}$, that is, if the coverage of agent $i_1$ starts before the coverage of agent $i_2$. When this happens, the binary variable $c^{cc}_{i_1.j_1,i_2,j_2} = 1$.
Since $i_1$ starts covering earlier, we have to assure that its coverage is not split by the end of the period, that is, its departure time is not translated to the first part of the period, as for the coverage of $\mathbf{q}_3$ by $i_3$ in Fig.~\ref{Subfig:Scheduling}. Eq.~\eqref{12a_e} guarantees this.
In the third place, Eq.~\eqref{12a_a} assures that the coverage of agent $i_2$ starts after the coverage of $i_1$ has ended, i.e., $D_{i_1,j_1} \leq A_{i_2,j_2}$.
The last restriction guarantees that, if the coverage of $i_2$ has to be split, i.e., $c^d_{i_2,j_2} - c^a_{i_1,j_1} = 1$, it finishes before the coverage of $i_1$ starts.

The second group of restrictions is
\begin{subequations}
\label{Eq:Constr_3_2}
\begin{align}
    A_{i_2,j_2} - A_{i_1,j_1} & \leq R \, c^{cc}_{i_1.j_1,i_2,j_2}, \label{12b_d} \\
    A_{i_2,j_2} - D_{i_2,j_2} & \leq R \,  c^{cc}_{i_1.j_1,i_2,j_2}, \label{12b_e} \\
    D_{i_2,j_2} - A_{i_1,j_1} & \leq -\varepsilon + R \,  c^{cc}_{i_1.j_1,i_2,j_2}, \label{12b_a}\\
    D_{i_1,j_1} - A_{i_2,j_2} & \leq -\varepsilon + R \, c^{cc}_{i_1.j_1,i_2,j_2} \nonumber \\
    & \quad + R \, (1 - c^d_{i_1,j_1} + c^a_{i_1,j_1}). \label{12b_c}
\end{align}
\end{subequations}
Opposite to the first one, this second group is activated when $A_{i_2,j_2} \leq A_{i_1,j_1}$ or, equivalently, when $c^{cc}_{i_1.j_1,i_2,j_2} = 0$.

\subsection{Collision Avoidance During Motion}

The second situation in which a collision between a pair of agents may occur is when both of them are moving or one is moving and the other is covering a point.
This can be prevented in the same way that the collisions during coverage, by avoiding the overlap of the movements and coverages that may cause a conflict.
In fact, we propose the same restrictions as in~\eqref{Eq:Constr_3_1}-\eqref{Eq:Constr_3_2}  for the two cases: the movements of two agents, and a movement of an agent and a coverage of a point by another agent.

We define as $\delta_{i,j}$ and $\alpha_{i,j}$ the initial and final times of the movement  between $\Gamma_i(j)$ and $\Gamma_i(j+1)$, respectively.
They are defined opposite to the coverage times since $\delta_{i,j} = d_{i,j}$ is the departure of the movement and $\alpha_{i,j} = a_{i,j+1}, j \in \{1,\dots,|\Gamma_i|-1\},$ is the arrival of the movement.
For $j = |\Gamma_i|$, we have $\alpha_{i,|\Gamma_i|} = 1$.
Similarly to the coverage times, we can express the departure and arrival times of the movements in the team plan as a function of the times in the agent period and the binary variables $c^a_{i,j}$ and $c^d_{i,j}$ as follows:
\begin{subequations}
\label{Eq:MoveTimesTeamPeriod}
\begin{align}
\Delta_{i,j} & = \varphi_i + \delta_{i,j} - c^d_{i,j}, \\
\Lambda_{i,j} & = \varphi_i + \alpha_{i,j} - c^a_{i,j+1}.
\end{align}
\end{subequations}
For $j = |\Gamma_i|$, we define the binary variable $c^a_i$ that represents if the final time of the last movement of each agent is greater than one or not. It requires the following restrictions:
\begin{align*}
\varphi_i + \alpha_{i,|\Gamma_i|} & \leq 1 + R \, c^a_i, \\
-(\varphi_i + \alpha_{i,|\Gamma_i|}) & \leq -1 + R \, ( 1 - c^a_i), \\
c^d_{i,|\Gamma_i|} - c^a_i & \leq 0.
\end{align*}
    
In order to decide if two movements can lead to a collision, we calculate the minimum distance between the two trajectories of the movements, $d^{mm}_{i_1.j_1,i_2,j_2}$ and determine that a collision is possible if such distance is lower than the sum of the sizes of the agents, i.e., $d^{mm}_{i_1.j_1,i_2,j_2} < r_{i_1} + r_{i_2}$.
If the pair of movements may result into conflict, the following two sets of constraints are included in the problem. These sets are the same as~\eqref{Eq:Constr_3_1}-\eqref{Eq:Constr_3_2}, respectively, but for the departure and arrival times of the movements.
The first set,
\begin{subequations}
\label{Eq:Constr_521}
\begin{align}
    \Delta_{i_1,j_1} - \Delta_{i_2,j_2} & \leq R \, (1 - c^{mm}_{i_1.j_1,i_2,j_2}), \label{521a} \\
    \Delta_{i_1,j_1} - \Lambda_{i_1,j_1} & \leq R \, (1 - c^{mm}_{i_1.j_1,i_2,j_2}), \label{521b} \\
    \Lambda_{i_1,j_1} - \Delta_{i_2,j_2} & \leq -\varepsilon + R \, (1 - c^{mm}_{i_1.j_1,i_2,j_2}), \label{521c} \\
    \Lambda_{i_2,j_2} - \Delta_{i_1,j_1} & \leq -\varepsilon + R \, (1 - c^{mm}_{i_1.j_1,i_2,j_2}) \nonumber \\
    & \quad + R \, (1 - c^d_{i_2,j_2} + c^a_{i_2,j_2+1}), \label{521d}
\end{align}
\end{subequations}
is active when the movement of agent $i_1$ starts before the movement of $i_2$, i.e., $\Delta_{i_1,j_1} \leq \Delta_{i_2,j_2}$ in Eq.~\eqref{521a}. This is represented by the binary variable $c^{mm}_{i_1.j_1,i_2,j_2} = 1$. The second constraint represents that the movement of agent $i_1$ cannot be split by the end of the period; the third one, that the movement of $i_2$ must start after the movement of $i_1$ has ended; and the last one that, if the movement of $i_2$ is split by the end of the period, it must finish before $i_1$ starts moving.

Equivalently, the second set of constraints activates when the movement of agent $i_2$ starts before the movement of $i_1$:
\begin{subequations}
\label{Eq:Constr_522}
\begin{align}
    \Delta_{i_2,j_2} - \Delta_{i_1,j_1} & \leq R \, c^{mm}_{i_1.j_1,i_2,j_2}, \label{522a} \\
    \Delta_{i_2,j_2} - \Lambda_{i_2,j_2} & \leq R \, c^{mm}_{i_1.j_1,i_2,j_2}, \label{522b} \\
    \Lambda_{i_2,j_2} - \Delta_{i_1,j_1} & \leq -\varepsilon + R \, c^{mm}_{i_1.j_1,i_2,j_2}, \label{522c}\\
    \Lambda_{i_1,j_1} - \Delta_{i_2,j_2} & \leq -\varepsilon + R \, c^{mm}_{i_1.j_1,i_2,j_2} \nonumber \\
    & \quad + R \, (1 - c^d_{i_1,j_1} + c^a_{i_1,j_1+1}), \label{522d}
\end{align}
\end{subequations}
and the meaning is the same as~\eqref{Eq:Constr_521} replacing $i_1$ by $i_2$ and vice versa.

The second type of collisions during the motion of an agent $i_1$ is with another agent $i_2$ that is covering a point. In that case, we calculate the minimum distance between the trajectory of the movement of $i_1$ and the point where $i_2$ is covering, $d^{mc}_{i_1.j_1,i_2,j_2}$.
If a collision may happen, i.e., $d^{mc}_{i_1.j_1,i_2,j_2} < r_{i_1} + r_{i_2}$, we include the same two sets of constraints as before with the departure and arrival times of the movement of $i_1$ and the arrival and departure times of  $i_2$ to the point that it must cover.
The first set,
\begin{align*}
    \Delta_{i_1,j_1} - A_{i_2,j_2} & \leq R \, (1 - c^{mc}_{i_1.j_1,i_2,j_2}),  \\
    \Delta_{i_1,j_1} - \Lambda_{i_1,j_1} & \leq R \, (1 - c^{mc}_{i_1.j_1,i_2,j_2}), \\
    \Lambda_{i_1,j_1} - A_{i_2,j_2} & \leq -\varepsilon + R \, (1 - c^{mc}_{i_1.j_1,i_2,j_2}), \\
    D_{i_2,j_2} - \Delta_{i_1,j_1} & \leq -\varepsilon + R \, (1 - c^{mc}_{i_1.j_1,i_2,j_2}) \\
    & \quad + R \, (1 + c^d_{i_2,j_2} - c^a_{i_2,j_2}),
\end{align*}
is active if the movement starts before the coverage and the second,
\begin{align*}
    A_{i_2,j_2} - \Delta_{i_1,j_1} & \leq R \,  c^{mc}_{i_1.j_1,i_2,j_2}, \\
    A_{i_2,j_2} - D_{i_2,j_2} & \leq R \, c^{mc}_{i_1.j_1,i_2,j_2}, \\
    D_{i_2,j_2} - \Delta_{i_1,j_1} & \leq -\varepsilon + R \,  c^{mc}_{i_1.j_1,i_2,j_2}, \\
    \Lambda_{i_1,j_1} - A_{i_2,j_2} & \leq -\varepsilon + R \,  c^{mc}_{i_1.j_1,i_2,j_2} \\
    & \quad + R \, (1 - c^d_{i_1,j_1} + c^a_{i_1,j_1+1}),
\end{align*}
if the opposite happens.
The interpretation of these constraint sets is the same as for~\eqref{Eq:Constr_3_1}-\eqref{Eq:Constr_3_2} and~\eqref{Eq:Constr_521}-\eqref{Eq:Constr_522}.

\subsection{Cost Function}

The proposed restrictions guarantee that in the resulting schedule no collisions between agents occur at any time.
In some applications, finding such solution may be enough and the problem can be posed as a Constraint Satisfaction Problem.
Nevertheless, in many other applications, it is desirable to find a solution that is not only feasible but also optimizes some kind of metric.
In particular, we aim to minimize the time in which two or more agents are moving simultaneously:
\begin{equation}
    f_{schedule} = \sum_{i_1=1}^{I-1} \sum_{i_2=i_1+1}^{I} \max\big( 0, \min(\Lambda_{i_{k_1},j_{k_1}} - \Delta_{i_{k_2},j_{k_2}}) \big),
    \label{Eq:CostFunction}
\end{equation}
with $k_1,k_2 = 1,2$. $f_{schedule}$ represents the sum of the times in which each pair of movements of different agents are overlapped.
This function is motivated by the mobile induction application, that we introduce in Section~\ref{sec:experiments}, where the cost function intends to minimize the changes in power requested from the electric grid. Although the cost could also be expressed as a restriction if only one agent were allowed to move at the same time, it is more appropriate to define it as we propose to yield a feasible solution that minimizes simultaneous motion.

The problem of minimizing~\eqref{Eq:CostFunction} can be transformed to the standard MILP formulation as follows.
First, we introduce the auxiliary variables $x_{i_1,j_1,i_2,j_2}$ to be greater or equal to the minimum inside~\eqref{Eq:CostFunction}, that is,
\begin{equation}
    f_{schedule} = \sum_{i_1=1}^{I-1} \sum_{i_2=i_1+1}^{I} \max( 0, x_{i_1,j_1,i_2,j_2} ),
    \label{Eq:CostFunction_1}
\end{equation}
with
\begin{align*}
    x_{i_1,j_1,i_2,j_2} \geq \min( &\Lambda_{i_1,j_1} - \Delta_{i_1,j_1}, \Lambda_{i_1,j_1} - \Delta_{i_2,j_2}, \\
    &\Lambda_{i_2,j_2} - \Delta_{i_1,j_1}, \Lambda_{i_2,j_2} - \Delta_{i_2,j_2} ).
\end{align*}
This can be reduced to one of the following constraints
\begin{subequations}
\label{Eq:CostRestr_x}
\begin{align}
    x_{i_1,j_1,i_2,j_2} & \geq \Lambda_{i_1,j_1} - \Delta_{i_1,j_1} - R \, e_{i_1,j_1,i_2,j_2}^1, \label{Subeq:CostRestr_4_1} \\
    x_{i_1,j_1,i_2,j_2} & \geq \Lambda_{i_1,j_1} - \Delta_{i_2,j_2} - R \, e_{i_1,j_1,i_2,j_2}^2, \label{Subeq:CostRestr_4_2}  \\
    x_{i_1,j_1,i_2,j_2} & \geq \Lambda_{i_2,j_2} - \Delta_{i_1,j_1} - R \, e_{i_1,j_1,i_2,j_2}^3, \label{Subeq:CostRestr_4_3}  \\
    x_{i_1,j_1,i_2,j_2} & \geq \Lambda_{i_2,j_2} - \Delta_{i_2,j_2} - R \, e_{i_1,j_1,i_2,j_2}^4, \label{Subeq:CostRestr_4_4}
\end{align}
\begin{equation}
    e_{i_1,j_1,i_2,j_2}^1 + e_{i_1,j_1,i_2,j_2}^2 + e_{i_1,j_1,i_2,j_2}^3 + e_{i_1,j_1,i_2,j_2}^4 = 3,
    \label{Subeq:CostRestr_4_0}
\end{equation}
\end{subequations}
where $e_{i_1,j_1,i_2,j_2}^k, k \in \{1,\dots,4\}$, are binary variables to activate the constraint between~\eqref{Subeq:CostRestr_4_1} and~\eqref{Subeq:CostRestr_4_4} that gives the minimum value of $x_{i_1,j_1,i_2,j_2}$, thanks to Eq.~\eqref{Subeq:CostRestr_4_0}.
Note that~\eqref{Subeq:CostRestr_4_1}-\eqref{Subeq:CostRestr_4_4} depend on $c^a_{i,j}$ and $c^d_{i,j}$ through~\eqref{Eq:MoveTimesTeamPeriod} and the difference $\Lambda_{i_{k_1},j_{k_1}} - \Delta_{i_{k_2},j_{k_2}}$ may be different depending on their values, i.e., if any of the movements is split between the end and the beginning of the team plan. For the sake of readability we develop a more detailed description of these constraints in Appendix~\ref{App:CostRestr}.

Second, we introduce the auxiliary variables $z_{i_1,j_1,i_2,j_2}$ to be greater or equal to the maximum inside~\eqref{Eq:CostFunction}:
\begin{equation}
    f_{schedule} = \sum_{i_1=1}^{I-1} \sum_{i_2=i_1+1}^{I} z_{i_1,j_1,i_2,j_2},
    \label{Eq:CostFunction_2}
\end{equation}
such that
\begin{subequations}
\label{Eq:CostRestr_z}
\begin{align}
    z_{i_1,j_1,i_2,j_2} & \geq 0, \\
    z_{i_1,j_1,i_2,j_2} & \geq x_{i_1,j_1,i_2,j_2}.
\end{align}
\end{subequations}
Since the objective~\eqref{Eq:CostFunction_2} is to minimize the sum of $z_{i_1,j_1,i_2,j_2}$, it can be seen that $x_{i_1,j_1,i_2,j_2}$ will also be minimized by activating only the proper constraint in~\eqref{Subeq:CostRestr_4_1}-\eqref{Subeq:CostRestr_4_4}.

\subsection{Optimal Schedule}
\label{Subsec:OptimSched}

Finally, we are in position to formulate the complete problem of finding the optimal schedule for the team of agents.

\begin{problem}
\label{Prob:Sched}
The optimal periodic schedule, in which each agent executes its own periodic plan and which guarantees that no collisions occur and that the time in which two or more movements overlap is minimum, is the solution of the following MILP:
\begin{equation}
    \min f_{schedule}
\end{equation}
subject to the restrictions introduced between~\eqref{Eq:Constr_1_12} and \eqref{Eq:CostRestr_z}.
\end{problem}

The variables that define the team schedule are $\varphi_i$, which represent the time that each individual plan has to be shifted to produce the team plan. Recall that only $I-1$ of these variables are needed.
It is important to emphasize that, although the number of additional variables and restrictions is high, the problem can be solved efficiently using standard state-of-the-art solvers~\cite{cplex}.
We provide more details on the computational time in the simulations, Section~\ref{sec:simulations}.
In addition, the solution to the problem is independent of the value of the period since all the times are obtained as a fraction of such period. In fact, the period can be calculated separately depending on the desired performance of the system. For instance, it can be set to the maximum time that a point can remain uncovered or calculated depending on the maximum time allocated to move.

On the downside, there are several reasons for which a feasible solution may not exist. For instance, it may happen if two agents have a common path but in opposite directions, if the space in which the agents can move is limited, or if an agent has to go through a point of the environment that is never left unoccupied by another agent.
They can be avoided by modifying the paths of the agents in at least three ways: (i) invert the direction of the movement, (ii) change the points that are assigned to each agent, or (iii) use a different cost function for the path optimization~\eqref{Eq:QuadProgram}.
The best solution in each case depends on the application.

\section{Simulation Results}
\label{sec:simulations}

In this section we evaluate in simulation the results and computational times of our solution for different cost functions used in the calculation of the optimal times and productions. This function has an important influence in the final paths of the agents and, therefore, in the team plan and the quality of the coverage provided.
In particular, we focus on the case of linear functions because they simplify the problem and are appropriate for the application of mobile induction heating, that we introduce in Section~\ref{sec:experiments}.
The first function that we evaluate is
 \begin{equation*}
 f_1(\theta_{\mathbf{q},r}) = \sum\limits_{i \in \mathcal{I}} \sum\limits_{j \in \Gamma_i} \left( -\theta_{i,j} + \frac{1}{\rho_i^{\max}} \, \rho_{i,j} \right),
 \label{Eq:CostFuncTimesPowers}
 \end{equation*}
 in which the goal is to maximize the assigned times and minimize the normalized productions to provide a coverage as homogeneous in time as possible.
The second alternative only tries to minimize the productions in order to reduce the maximum actions of the agents: 
 \begin{equation*}
 f_2(\theta_{\mathbf{q},r}) = \sum\limits_{i \in \mathcal{I}} \sum\limits_{j \in \Gamma_i} \frac{1}{\rho_i^{\max}} \, \rho_{i,j},
 \label{Eq:CostFuncPowers}
 \end{equation*}
In the third function we include a weight to the times with respect to $f_1$ that is the inverse of the distance from the agent to the point, normalized by the maximum distance that the agent can reach,
 \begin{equation*}
 f_3(\theta_{\mathbf{q},r}) = \sum\limits_{i \in \mathcal{I}} \sum\limits_{j \in \Gamma_i} \left( -\frac{1}{d_{\mathbf{p}_i^0,\mathbf{q}}} \, \theta_{i,j} + \frac{1}{\rho_i^{\max}} \, \rho_{i,j} \right),
 \label{Eq:CostFuncDistTimesPowers}
 \end{equation*}
and in the fourth one we also weight the productions with the relative importance of the required coverage $P^*(\mathbf{q})$ with respect to the maximum production of the agent $\rho_i^{\max}$.
 \begin{equation*}
 f_4(\theta_{\mathbf{q},r}) = \sum\limits_{i \in \mathcal{I}} \sum\limits_{j \in \Gamma_i} \left( -\frac{1}{d_{\mathbf{p}_i^0,\mathbf{q}}} \, \theta_{i,j} + \frac{P^*(\mathbf{q})}{{\rho_i^{\max}}^2} \, \rho_{i,j} \right),
 \label{Eq:CostFuncDistTimesWeightedPowers}
 \end{equation*}
In the last two alternatives the times are weighted using the clustering procedure introduced in~\cite{palacios2016times}, without the productions in
 \begin{equation*}
 f_5(\theta_{\mathbf{q},r}) = \sum\limits_{i \in \mathcal{I}} \sum\limits_{j \in \Gamma_i} - \omega_{i,j} \, \theta_{i,j},
 \label{Eq:CostFuncClustering}
 \end{equation*}
and with weighted productions in
 \begin{equation*}
 f_6(\theta_{\mathbf{q},r}) = \sum\limits_{i \in \mathcal{I}} \sum\limits_{j \in \Gamma_i} \left( - \omega_{i,j} \, \theta_{i,j} + \frac{P^*(\mathbf{q})}{{\rho_i^{\max}}^2} \, \rho_{i,j} \right).
 \label{Eq:CostFuncClusteringPowers}
 \end{equation*}
 
The evaluation consisted in 10 runs of the algorithm with different positions of the points for different numbers of agents and points, all of them for the six cost functions.
The required coverage was randomly selected between 400 and 2200 units while the maximum production of the agents was 5000 units.
The number of agents and points was selected in a way that the team had enough production to cover the points. This restricts to at most twice more points than agents.
 
\begin{figure}[!ht]
        \centering
                \includegraphics[width=0.3\textwidth]{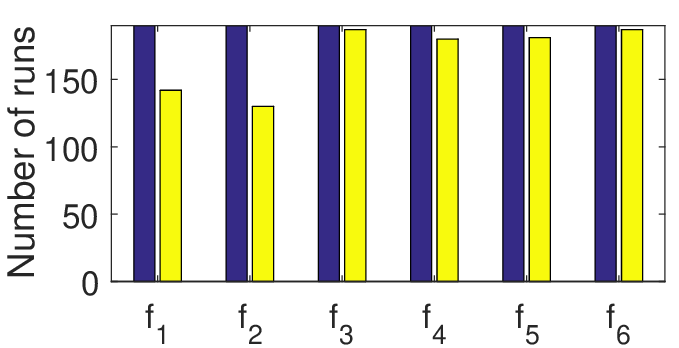}
    \caption{Number of solutions found for the times and productions (dark blue) and for the scheduling (yellow).}
    \label{Fig:NSolutions}
 
\end{figure}

Since the cost function influences the points that each agent finally visits, we first evaluate in how many cases the optimal solution was found, both for the calculation of times and productions and for the team plan scheduling.
In Fig.~\ref{Fig:NSolutions} it can be seen that a solution was found in all the 190 runs for the times and production. However, the first two cost functions only allowed a solution for the scheduling in the $75\%$ and $68\%$ of the cases while the others allowed more than $95\%$ of solutions. This supports the discussion on the existence of solution from in Section~\ref{Subsec:OptimSched}.

\begin{figure*}[!ht]
        \centering
        \subfloat[$f_1$.]{
                \includegraphics[height=2.74cm]{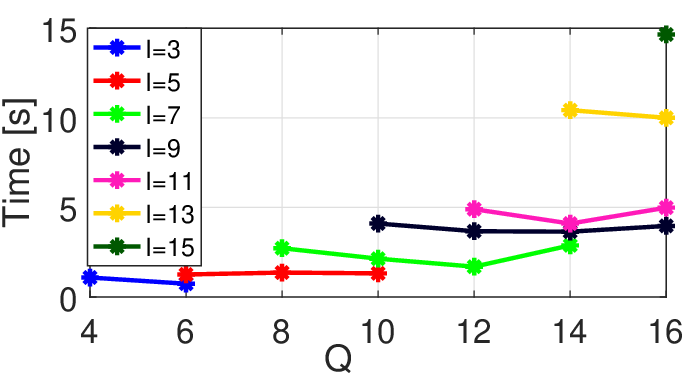}
                \label{Subfig:f_1}
                }
                \hfil
        \subfloat[$f_2$.]{
                \includegraphics[height=2.7cm]{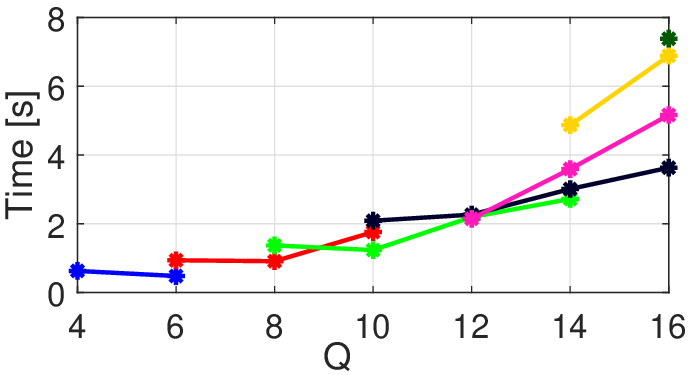}
                \label{Subfig:f_2}
                }
                \hfil
        \subfloat[$f_3$.]{
                \includegraphics[height=2.7cm]{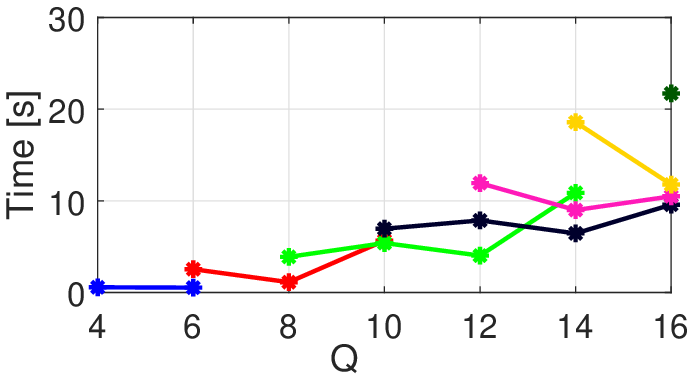}
                \label{Subfig:f_3}
                }
                
        \subfloat[$f_4$.]{
                \includegraphics[height=2.7cm]{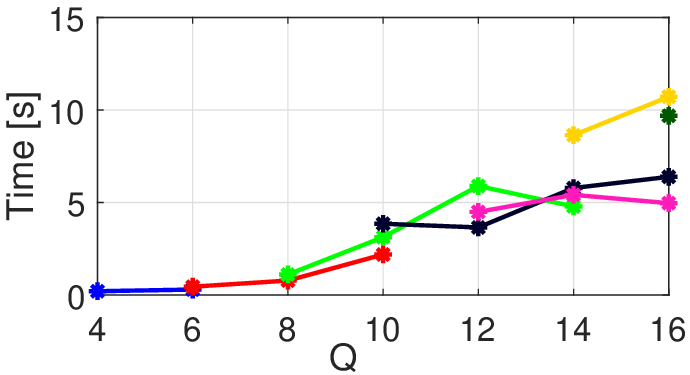}
                \label{Subfig:f_4}
                }
                \hfil
        \subfloat[$f_5$.]{
                \includegraphics[height=2.7cm]{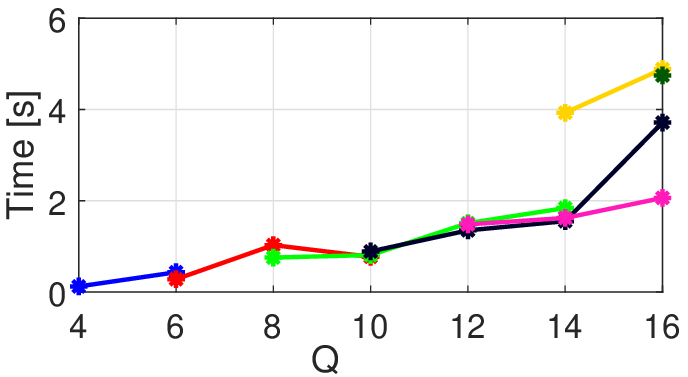}
                \label{Subfig:f_5}
                }
                \hfil
        \subfloat[$f_6$.]{
                \includegraphics[height=2.7cm]{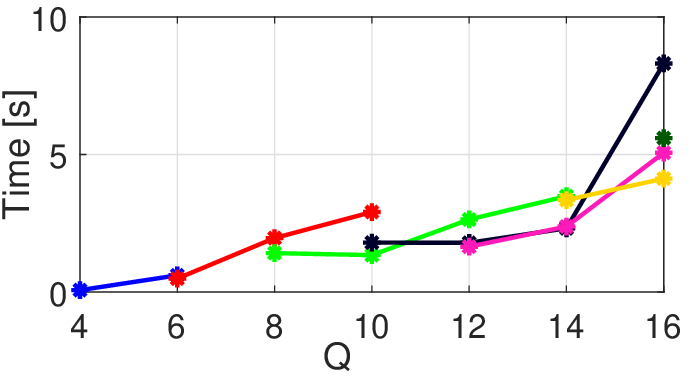}
                \label{Subfig:f_6}
                }
                \hfil
    \caption{Mean calculation time of the coverage times and productions solving problem~\eqref{Eq:QuadProgram}. Each figure represents the results with a different cost function and different colors represent different number of agents.}
    \label{Fig:MeanAssignmentTimes}

\end{figure*}

In Fig.~\ref{Fig:MeanAssignmentTimes} we show the computation times to obtain the coverage times and productions solving problem~\eqref{Eq:QuadProgram}.
Although in all the cases the increase of the time is more than linear with the number of agents and points, for a team of 13 or 15 agents, the computation requires around 20 seconds in the worst case, i.e., with $f_3$. This time is negligible for most  persistent coverage applications in which the agents are required to carry out the task indefinitely.
In addition, the difference between cost functions is significant. The worst-case computation required with $f_3$ is around 5 times lower with $f_5$ and in some cases, such as with $f_1$, the computation increases only with the number of robots.

\begin{figure}[!ht]
        \centering
                \includegraphics[width=0.3\textwidth]{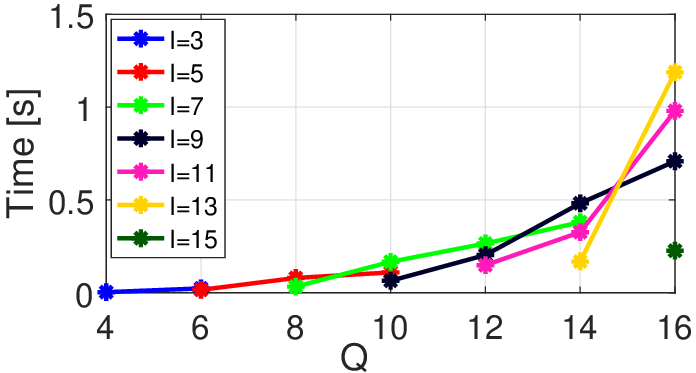}
    \caption{Mean calculation time of the scheduling solving Problem~\ref{Prob:Sched}.}
    \label{Fig:MeanSchedTimes}

\end{figure}

The results on the computation time of the scheduling are very similar for all the cost function and, therefore, we only show in Fig.~\ref{Fig:MeanSchedTimes} the times for $f_1$. 
These times are in the order of milliseconds for small teams and only increase to approximately one second for the biggest teams and numbers of points.
According to these results, the computational complexity resides in problem~\eqref{Eq:QuadProgram} rather than in Problem~\ref{Prob:Sched}.
Moreover, the small computation time with respect to potential period values supports the recalculation of the solution when $P^*(\mathbf{q})$ changes.

Finally we compare in Table~\ref{Tab:Metrics} six different metrics obtained from the simulations to give an idea of the behavior with each function.
The first metric is the average number of iterations of Algorithm 1 in~\cite{palacios2016times} to find the final solution. With $f_3$ and $f_5$ it is slightly bigger, which means that these cost functions allow the paths to be progressively shortened more than the others.
The second metric is the average number of movements per agent during each period, which is really similar for all the functions.
The third row includes the average time that each agent dedicates to cover the point in a period. The differences are again small with $f_5$ and $f_6$ producing only $2\%$-better results.
In the fourth place we calculate the maximum time that a point remains uncovered every period. The same functions $f_5$ and $f_6$ give better results even though there is no much difference between all of them.
The fifth metric shows the maximum production required to the agents and normalized by the maximum production available. In this case, $f_3$ performs much worse than the others while $f_4$ and $f_6$ require the smallest maximum productions.
Finally we compute a metric that determines the homogeneity of the coverage. It is defined as
 \begin{equation*}
 h = \sum\limits_{\mathbf{q} \in \mathcal{Q}} \sum\limits_{i \in \mathcal{I}}  \left( \theta_{i,j} - \frac{P^*(\mathbf{q})}{\rho_i^{\max}} \right)^2,
 \label{Eq:Homogeneity}
 \end{equation*}
and the results show that, on average, the cost functions $f_1$ and $f_2$ provide the most homogeneous coverage.
As a conclusion, one can see that all the alternatives have advantages and disadvantages and the selection must be made depending on the application. 
\begin{table}
\caption{Comparison of metrics between cost functions.}
\centering
\begin{tabular}{ccccccc}
\hline 
\hline 
 & $f_1$ & $f_2$ & $f_3$ & $f_4$ & $f_5$ & $f_6$ \\
\hline  
Iterations & 1.11 & \textbf{1.10} & 1.27 & 1.12 & 1.36 & 1.13 \\
Movements & \textbf{1.43} & 1.44 & 1.44 & 1.44 & 1.45 & 1.44 \\ 
Total Coverage & 0.95 & 0.95 & 0.94 & 0.95 & \textbf{0.97} & \textbf{0.97} \\ 
Uncovered Time &  0.30 & 0.30 & 0.32 & 0.31 & \textbf{0.29} & \textbf{0.29} \\
Max Production &  0.73 & 0.73 & 0.93 &  0.69 & 0.81 & \textbf{0.68} \\ 
Homogeneity & 3.03 & \textbf{2.93} & 3.18 & 3.16 & 3.45 & 3.24 \\ 
\hline 
\hline 
\label{Tab:Metrics}
\end{tabular}
\end{table}

\section{Application to Mobile Induction Heating}
\label{sec:experiments}

\begin{figure}[!ht]
        \centering
                \includegraphics[height=3.5cm]{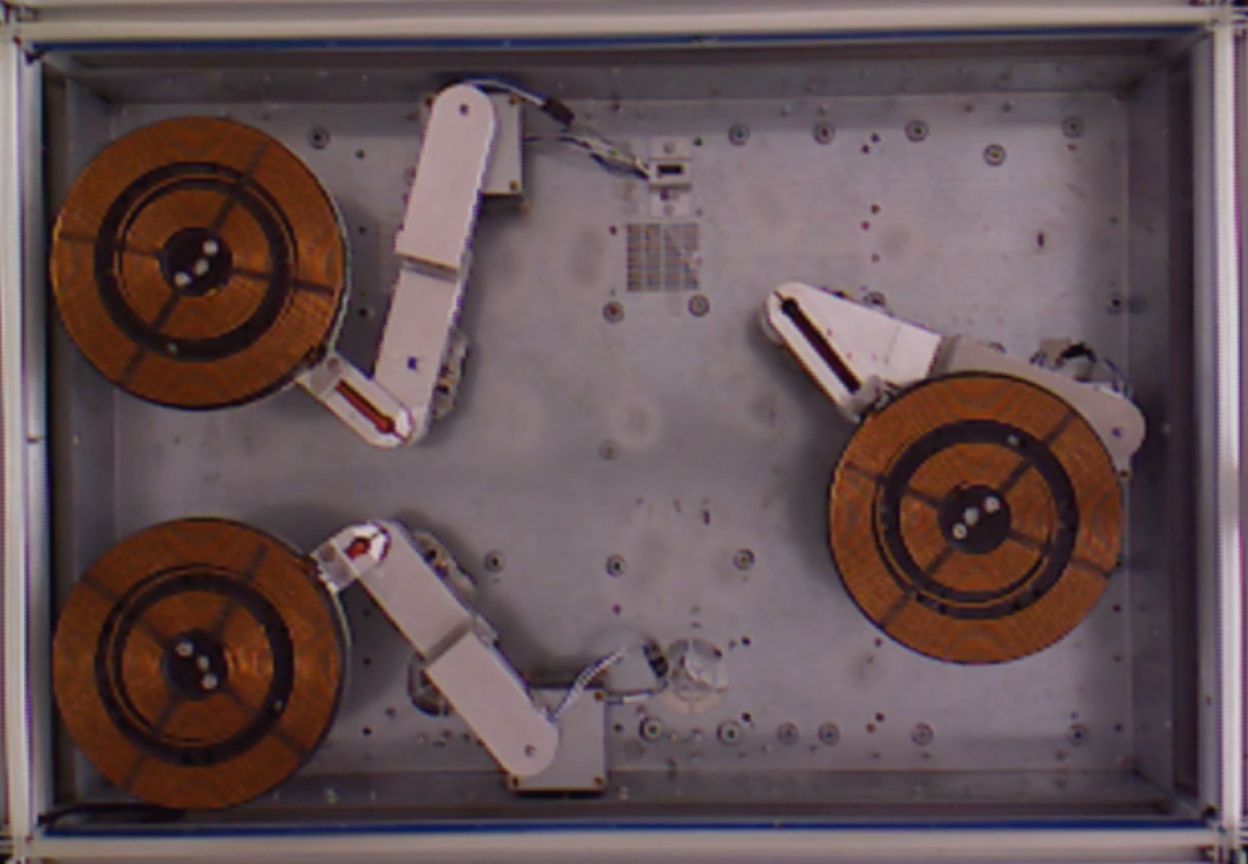}
    \caption{Prototype of induction hob with mobile inductors.}
    \label{Fig:ExperimentSnapshot}
\end{figure}

Throughout the paper we have presented a solution to persistent coverage with a team of generic agents.
Now we particularize the proposal to the application of induction heating in a domestic hob for cooking.
In this application, a finite set of cooking pots have to be persistently heated to maintain their temperature or received power as close as possible to the reference determined by the user.
This is achieved by means of induction technology with a group of inductors that are capable of transferring power to the pots and can move inside the hob.
In particular, we apply our solution to the prototype shown in Fig.~\ref{Fig:ExperimentSnapshot}.

The size of the prototype is 726 $\times$ 482 mm and it has three inductors of 180 mm of diameter.
Each one is attached to a robotic arm with two rotational joints.
This design allows two configurations of the arms in many positions of the inductors, one with the second joint to the left of the inductor and another with the joint to the right. However, when moving closer to the hob limits, only one configuration may be feasible without colliding with the border and to change between configurations the inductor must go through its home position.

The motion of the arms is given by stepper motors, which are controlled with Arduino boards and specific drivers.
The strategies are computed in Matlab and transmitted to the Arduino boards through the serial bus.
Since the interest of this validation is in the movements of the inductors and the periodic strategy, in these experiments the pots have been simulated and the power has not been actually provided in order to accelerate the experimentation and guarantee the safety of the motion system.

In this context, the discrete points represent the pots, the coverage level is their received power and the agents are the inductors.
The coverage level deteriorates over time because the pots cool down if they are not heated.
The maximum production depends on the design of the inductor and the coupling pot-inductor and the inductors may not be able to reach all the pots due to the physically-limited reachability of the robotic arms.

The cost function that we have proposed for the team plan scheduling, that aims to minimize the time in which two or more agents are moving simultaneously, is motivated by this application as follows.
When an inductor starts moving from one pot to the following, it stops heating and, therefore, it stops requesting power from the electric grid. In the same way, when it reaches a new pot and starts heating, it starts requesting power from the grid. These changes in requested power are limited by the European norm UNE-EN 61000-3-3, 2013. In order to comply with this norm, we minimize the time of simultaneous movements so that, when an inductor stops (starts) requesting power, we can increase (decrease) the request of the others to compensate the total request of the system.

Given the particular features of the prototype and its spatial restrictions, there may appear cases in which the scheduling may not have a feasible solution due to the collision avoidance restrictions.
For instance, if an inductor is in charge of only one pot, i.e., it remains all the time under such pot, and another inductor has to heat a different pot sufficiently close to the first one, then the second inductor may not be capable of reaching it without colliding with the first one.
To solve these issues, geometric solutions have been implemented ad hoc and, since we focus on the general approach to periodic persistent coverage, it is out of the scope of this paper.

For this particular application we calculate the period as $T = \max_{i \in \mathcal{I}} t_i^m / \theta_{\max}^m $, where $t_i^m$ are the actual times that the robots need to traverse the path and $\theta_{\max}^m$ is the maximum portion of the period that we allow for the motion, in this case, $0.3$.

\subsection{Experimental Results}

We provide experimental results of the periodic heating strategy with the prototype introduced before.
The results obtained for different configurations of the pots are conceptually the same and, therefore, we only show an example.
\begin{figure}[!ht]
        \centering
                \includegraphics[height=3.5cm]{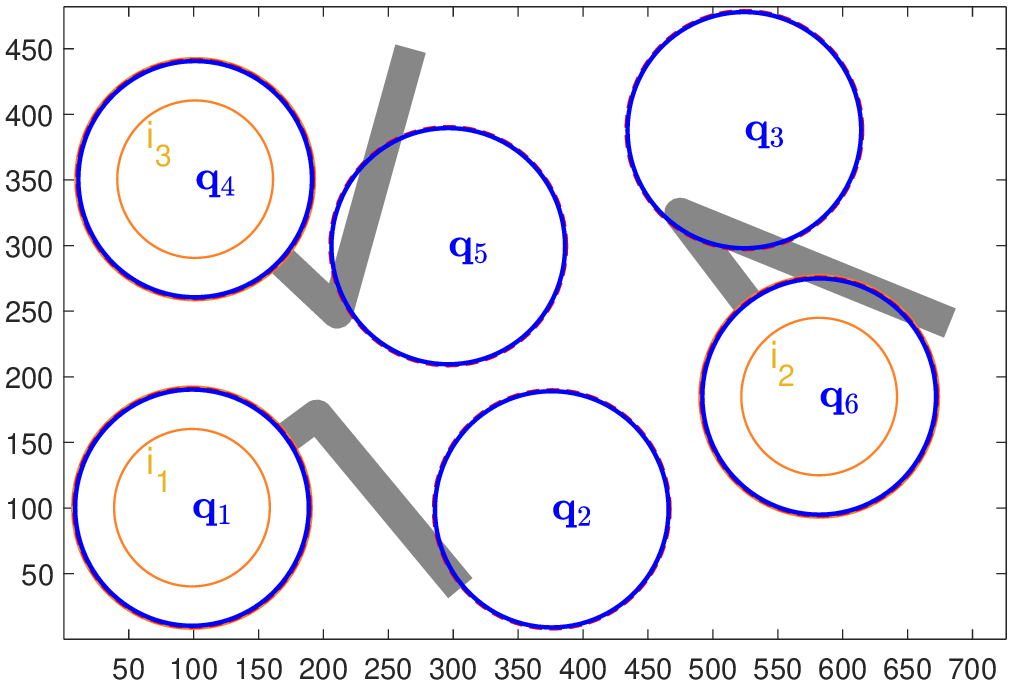}
    \caption{ Layout of one of the experiments. Blue circumferences represent the pots; orange circumferences, the two coils of the inductors; and grey segments, the two links of the robotic arms.}
    \label{Fig:ExperimentLayout}
\end{figure}

In Fig.~\ref{Fig:ExperimentLayout} the layout of the example is shown.
There are 6 virtual pots which are circles of diameter $18cm$. The pots require a constant $P^*(\mathbf{q})=800,1500,500,1500,2000$ and $1500W$, respectively.
At this particular time instant, the three inductors are heating their assigned pots.

\begin{figure}[!ht]
        \centering
                \includegraphics[width=0.3\textwidth]{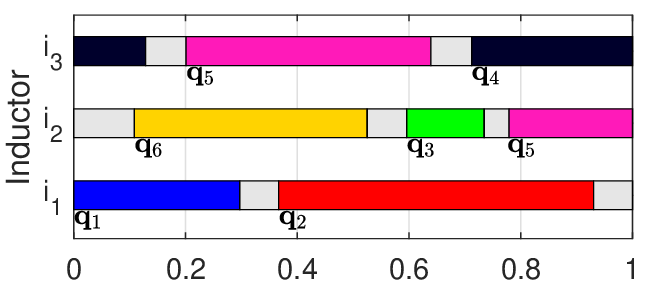}
    \caption{Optimal schedule obtained for the experiment.}
    \label{Fig:Schedule}
\end{figure}
\begin{figure*}[!ht]
        \centering
        \subfloat[Pot 1.]{
                \includegraphics[width=0.23\textwidth]{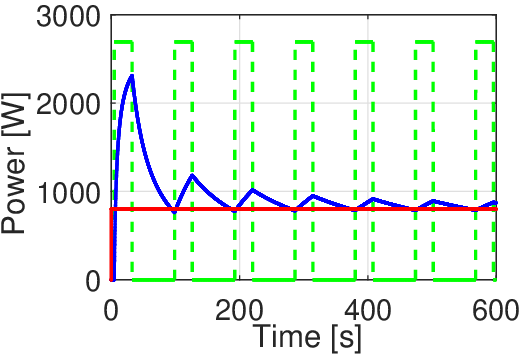}
                \label{Subfig:Pot1}
                }
                \hfil
        \subfloat[Pot 2.]{
                \includegraphics[width=0.23\textwidth]{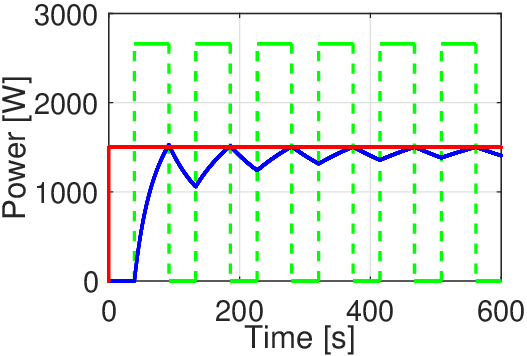}
                \label{Subfig:Pot2}
                }
                \hfil
        \subfloat[Pot 3.]{
                \includegraphics[width=0.23\textwidth]{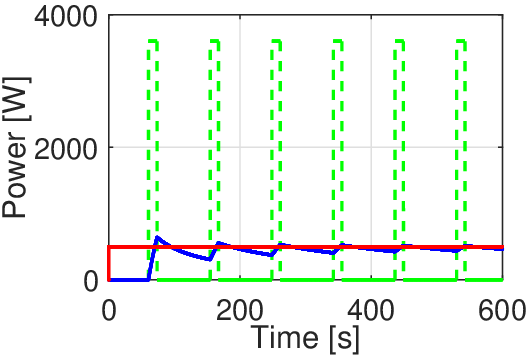}
                \label{Subfig:Pot3}
                }
                
        \subfloat[Pot 4.]{
                \includegraphics[width=0.23\textwidth]{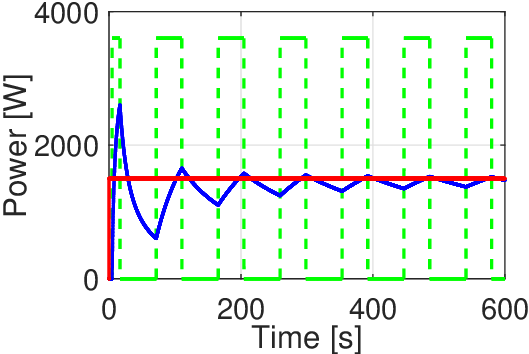}
                \label{Subfig:Pot4}
                }
                \hfil
        \subfloat[Pot 5.]{
                \includegraphics[width=0.23\textwidth]{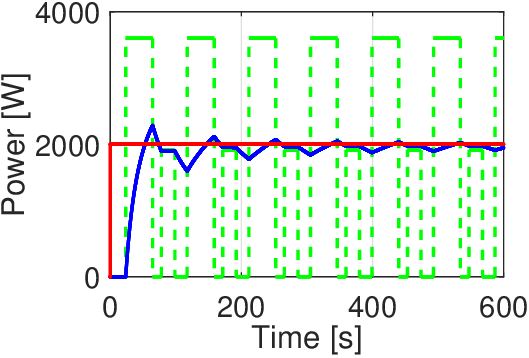}
                \label{Subfig:Pot5}
                }
                \hfil
        \subfloat[Pot 6.]{
                \includegraphics[width=0.23\textwidth]{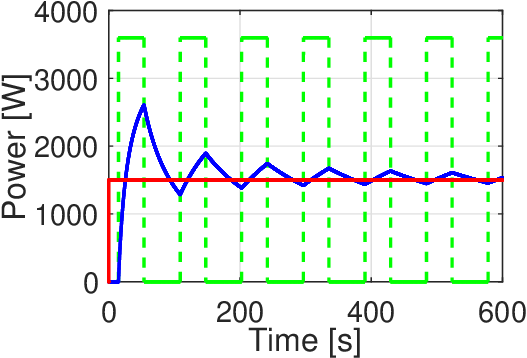}
                \label{Subfig:Pot6}
                }
    \caption{Power received by the pots during the experiment. The green line represents the instant received power; the red line, the requested power; and the blue line, the average received power.}
    \label{Fig:Pots}
\end{figure*}
The optimal schedule obtained for this configuration of inductors and pots is shown in Fig.~\ref{Fig:Schedule}. It can be seen that the inductor $i_2$, the right one in Fig.~\ref{Fig:ExperimentSnapshot}, has to spend most of its time heating pots $\mathbf{q}_3$ and $\mathbf{q}_6$ and also move to heat pot $\mathbf{q}_5$ in cooperation with inductor $i_3$, the top one, that is also in charge of pot $\mathbf{q}_4$. On the other hand, inductor $i_1$ alternatively heats pots $\mathbf{q}_1$ and $\mathbf{q}_2$.
 
The power received by the pots can be seen in Fig.~\ref{Subfig:Pot1}-~\ref{Subfig:Pot6}. One can see that all the pots receive power periodically for a period of time and then have to wait some time until the following coverage. This is a requirement of the system since there are more pots than inductors. However, the average received power quickly tends to the required power, meaning that, on average, the pots receive the desired power.

It is important to note that, although the curly power profile may result in important variations of the temperature of the pot, this effect can only be negative in boiling processes, where the bubbles can appear and disappear with the power changes. In other processes, such as deep frying or just warming, the variation of the temperature can hardly be appreciated by the user. Moreover, if a boiling process is taking place, the assignment of the inductors can be adjusted to keep an inductor under the boiling pot all the time while the other share the rest of the pots.

\begin{figure}[!ht]
        \centering \includegraphics[width=0.25\textwidth]{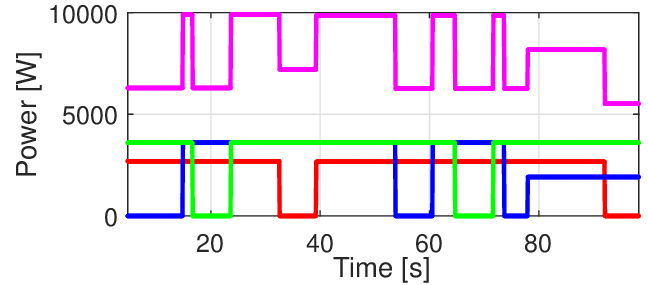}
    \caption{Power provided by the inductors during the experiment. The red, blue and green lines represent the power of each inductor and the magenta line, the total power.}
    \label{Fig:Powers}
\end{figure}

In Fig.~\ref{Fig:Powers} the power provided by the inductors is depicted along with the total power of the hob. The cost function of the scheduling, that minimizes the time in which more than one inductor is moving at the same time, favors that the variations of the total power are minimal. In addition, it allows us to adjust the total power of the hob when one inductor starts or stops requesting power with the other two inductors to satisfy the flicker norm.

Finally, we carried out an evaluation of all the possible solutions for this experiment using brute force. In particular, we calculated all the possible combinations of trajectories of the three inductors and tried to solve the second and fourth steps of Algorithm~\ref{Alg:Solution}. Out of the 1270 combinations, only 10 were feasible and the solution provided by our method was only a 1.41\% away from the best one in terms of coverage quality measured by $ \int_0^t \sum_{\mathbf{q}\in\mathcal{Q}} \big(Z^*(\mathbf{q},\tau) - Z(\mathbf{q},\tau) \big)^2 d\tau,$. Finding such best solution took 181s while our method only needed 0.95s.

\section{Conclusions}
\label{sec:conclusions}

In this paper we have introduced a periodic solution to the persistent coverage problem. We have proposed a path planning strategy for each individual agent and a quadratically constrained linear program to obtain the optimal coverage times and actions to satisfy the coverage objective of the environment. We have also presented an MILP to find a periodic team schedule in which no collisions occur.
Simulation results support our proposal and demonstrate that it is tractable for significant team and environment sizes. Moreover, we have shown experimental results for a heating application in which a set of pots is homogeneously heated in a domestic hob with mobile inductors.

\appendices

\section{Cost Function Constraints}
\label{App:CostRestr}

The constraints that arise from the minimum function in~\eqref{Eq:CostFunction} capture the duration of the overlap between each pair of movements.
In the case that one of the movements is completely overlapped with the other, the duration of the overlap is equal to the duration of such movement. Therefore, we can express Eq.~\eqref{Subeq:CostRestr_4_1} and~\eqref{Subeq:CostRestr_4_4} as
\begin{align*}
    x_{1,2} & \geq \Lambda_1 - \Delta_1 - R \, e_{1,2}^1, \\
    x_{1,2} & \geq \Lambda_2 - \Delta_2 - R \, e_{1,2}^4.
\end{align*}
In this appendix, we substitute subindices $i_1,j_1$ and $i_2,j_2$ by $1$ and $2$ for simplicity.

On the other hand, if the movements are only partially overlapped, the duration of such overlap is bounded by Eq.~\eqref{Subeq:CostRestr_4_2} or Eq.~\eqref{Subeq:CostRestr_4_3}.
However, the differences $\Lambda_1 - \Delta_2$ or $\Lambda_2 - \Delta_1$ may not represent such duration if one or both movements are split between the end and the beginning of the team plan, that happens if $c^d_k \equiv c^d_{i_k,j_k} = 0$ and $c^a_i \equiv c^a_{i_k,j_k+1} = 1$.
In the following we formulate the restrictions for all the cases in which none, one or both movements are split.
Each constraint is conditionally activated, applying the big number method, depending on $c^d_k$ and $c^a_k$. In particular, if $c^a_k - c^d_k = 0$, the movement is not split, and its corresponding constraints activated, and if $1 - c^a_k + c^d_k = 0$, the movement is split.

We set these two restrictions for the case in which none of the movements is split:
\begin{align*}
    x_{1,2} & \geq \Lambda_1 - \Delta_2 - R \, e_{1,2}^1 - R \, ( c^a_1 - c^d_1 ) - R \, ( c^a_2 - c^d_2 ), \\
    x_{1,2} & \geq \Lambda_2 - \Delta_1 - R \, e_{1,2}^4 - R \, ( c^a_1 - c^d_1 ) - R \, ( c^a_2 - c^d_2 ) \label{Subeq:CostRestr_4_3_a}
\end{align*}

Similarly, when both are split, the constraints are
\begin{align*}
    x_{1,2} \geq \Lambda_1 - (\Delta_2-1) - R \, e_{1,2}^1 & - R ( 1 - c^a_1 - c^d_1 ) \nonumber \\
    & - R \, ( 1 - c^a_2 - c^d_2 ), \\
    x_{1,2} \geq \Lambda_2 - (\Delta_1-1) - R \, e_{1,2}^4 & - R ( 1 - c^a_1 - c^d_1 ) \nonumber \\ & - R \, ( 1 - c^a_2 - c^d_2 ).
\end{align*}
In this case we only subtract 1 to the initial times to overcome the split.

We proceed in a similar way when only one of the movements is split although in these cases it is slightly more complicated.
In the first place we focus in the case in which the movement of agent 1 is split.
We check which part of the movement may overlap with the movement of agent $i_2$. This is done by introducing a binary variable $b_{1,2}^1$ such that $b_{1,2}^1 = 1$ if $\Delta_1 \leq \Lambda_2$, meaning that the second part is overlapped, and $b_{1,2}^1 = 0$, otherwise, meaning that the candidate for overlap is the first part.
This variable is obtained from the following constraints:
\begin{align*}
    \Delta_1 - \Lambda_2 & \leq R \, ( 1 - c^a_1 - c^d_1 ) + R \, ( c^a_2 - c^d_2 ) + R \, (1 - b_{1,2}^1), \\
    \Lambda_2 - \Delta_1 & \leq R \, ( 1 - c^a_1 - c^d_1 ) + R \, ( c^a_2 - c^d_2 ) + R \, b_{1,2}^1.
\end{align*}
Depending on the value of $b_{1,2}^1$ we transform the constraints differently.
When $b_{1,2}^1 = 1$ we add one to the final time of the movement of $i_1$. This can be seen as moving the part of the movement that is at the beginning of the plan to the right side. Thus, the constraints are
\begin{align*}
    x_{1,2} & \geq (\Lambda_1 + 1) - \Delta_2 - R \, e_{1,2}^1 - R \, ( 1 - c^a_1 - c^d_1 ) \\
    & \quad - R \, ( c^a_2 - c^d_2 ) - R \, ( 1 - b_{1,2}^1 ), \\
    x_{1,2} & \geq \Lambda_2 - \Delta_1 - R \, e_{1,2}^4 - R \, ( 1 - c^a_1 - c^d_1 ) \\
    & \quad - R \, ( c^a_2 - c^d_2 ) - R \, ( 1 - b_{1,2}^1 ).
\end{align*}
On the contrary, when $b_{1,2}^1 = 0$, we subtract one to the initial time of the movement to compare the movement of agent $i_2$ with the left part of the movement of $i_1$. The constraints result in
\begin{align*}
    x_{1,2} & \geq \Lambda_1 - \Delta_2 - R \, e_{1,2}^1 - R \, ( 1 - c^a_1 - c^d_1 ) \\
    & \quad - R \, ( c^a_2 - c^d_2 ) - R \, b_{1,2}^1, \\
    x_{1,2} & \geq \Lambda_2 - (\Delta_1-1) - R \, e_{1,2}^4 - R \, ( 1 - c^a_1 - c^d_1 ) \\
    & \quad - R \, ( c^a_2 - c^d_2 ) - R \, b_{1,2}^1.
\end{align*}

The same procedure can be followed for the case in which the movement of $i_2$ is split.
We include the binary variable $b_{1,2}^2$ with
\begin{align*}
    \Delta_2 - \Lambda_1 & \leq R \, ( c^a_1 - c^d_1 ) + R \, ( 1 - c^a_2 - c^d_2 ) + R \, (1 - b_{1,2}^2), \\
    \Lambda_1 - \Delta_2 & \leq R \, ( c^a_1 - c^d_1 ) + R \, ( 1 - c^a_2 - c^d_2 ) + R \, b_{1,2}^2.
\end{align*}
When $b_{1,2}^2 = 1$, the constraints are
\begin{align*}
    x_{1,2} & \geq \Lambda_1 - \Delta_2 - R \, e_{1,2}^1 - R \, ( c^a_1 - c^d_1 ) \\
    & \quad - R \, ( 1 - c^a_2 - c^d_2 ) - R \, ( 1 - b_{1,2}^2 ), \\
    x_{1,2} & \geq (\Lambda_2 + 1) - \Delta_1 - R \, e_{1,2}^4 - R \, ( c^a_1 - c^d_1 ) \\
    & \quad - R \, ( 1 - c^a_2 - c^d_2 ) - R \, ( 1 - b_{1,2}^2 ),
\end{align*}
and, otherwise,
\begin{align*}
    x_{1,2} & \geq \Lambda_1 - (\Delta_2-1) - R \, e_{1,2}^1 - R \, ( c^a_1 - c^d_1 ) \\
    & \quad - R \, ( 1 - c^a_2 - c^d_2 ) - R \, b_{1,2}^2, \\
    x_{1,2} & \geq \Lambda_2 - \Delta_1 - R \, e_{1,2}^4 - R \, ( c^a_1 - c^d_1 ) \\
    & \quad - R \, ( 1 - c^a_2 - c^d_2 ) - R \, b_{1,2}^2.
\end{align*}

Finally, note that for the implementation and resolution of the problem, the constraints~\eqref{Subeq:CostRestr_4_1}-\eqref{Subeq:CostRestr_4_4} have to be substituted by all the restrictions of this appendix.

\bibliographystyle{IEEEtran}

\end{document}